\documentclass[letterpaper, 10 pt, conference]{ieeeconf}  

\IEEEoverridecommandlockouts                              

\usepackage[utf8]{inputenc}
\usepackage[T1]{fontenc}
\usepackage{url}

\usepackage{amssymb}  

\usepackage{xcolor,amsmath}
\usepackage{mathtools}
\usepackage[thinc]{esdiff}
\usepackage[italicdiff]{physics}

\usepackage{algorithm}
\usepackage{algpseudocode}
\usepackage{amsmath}  

\usepackage{booktabs} 

\usepackage{graphicx}
\usepackage[caption=false,font=footnotesize]{subfig}

\usepackage{tikz-cd}
\usetikzlibrary{shapes.geometric, arrows}
\usetikzlibrary{fit}

\usepackage{tcolorbox}
\tcbuselibrary{listings, skins, breakable}

\newcommand{\FriW}[1]{\emph{FriWalk}}

\newcommand{\rev}[1]{#1}

\newcommand{\pao}[1]{#1}

\begin{document}

\title{\LARGE \bf An Infrastructure-less, Control-Independent Solution
  to Relative Localisation of a Team of Mobile Robots using Ranging
  Measurements
}

\author{
    Paolo Golinelli~\IEEEmembership{Student~Member~IEEE},
    Tommaso Faraci~\IEEEmembership{Student~Member~IEEE}, \\
    Daniele Fontanelli~\IEEEmembership{Fellow~Member~IEEE}
    \thanks{
         P. Golinelli and D. Fontanelli are with the Department of Industrial Engineering, University of Trento, Trento, Italy.
         \{name.surname\}@unitn.it.
    }
    \thanks{
         T. Faraci is with the Department of Information Engineering and Computer Science, University of Trento, Trento, Italy. \{name.surname\}@unitn.it.
    }
}

\begin{minipage}{\textwidth}
This paper has been accepted for publication in the \textit{IEEE International Conference on Robotics and Automation (ICRA)}. The final version will be available via IEEE Xplore.

\vspace{2em}


\vspace{2em}

\copyright 2026 IEEE. Personal use of this material is permitted. Permission from IEEE must be obtained for all other uses, in any current or future media, including reprinting/republishing this material for advertising or promotional purposes, creating new collective works, for resale or redistribution to servers or lists, or reuse of any copyrighted component of this work in other works.
\end{minipage}

\markboth{IEEE International Conference on Robotics and Automation (ICRA). Accepted January 2026}
{Golinelli \MakeLowercase{\textit{et al.}}: An Infrastructure-less, Control-Independent Solution
  to Relative Localisation of a Team of Mobile Robots using Ranging Measurements}


\maketitle


\DeclareGraphicsExtensions{.eps}  


\begin{abstract}
The ability to localise teams of robots is essential for applications ranging from robotic fleets in unstructured environments to cooperative control and navigation tasks. In such contexts, fixed infrastructure is often unavailable, deployments must be fast and flexible, and system requirements must be minimal.  We present a decentralised cooperative localisation algorithm that addresses all these challenges at once. The method is anchor-less, fully decentralised, and, unlike most existing approaches, does not require controlling the robots motion to ensure team observability. It relies only on local odometry, sparse inter-agent ranging measurements, and short-range communication, all of which are widely available in practice.  The algorithm adopts a multi-hypothesis Bayesian framework that maintains the entire set of feasible solutions, ensuring robustness under transient unobservable conditions. Moreover, through information sharing, each agent benefits from the estimates of the entire group, even in partially connected conditions.
\end{abstract}


\section{INTRODUCTION} \label{sec:intro}

Many advanced robotic applications require reliable
localisation of large groups of agents~\cite{target_loc,blimp_loc}.
Tracking fleets of robots is not only of industrial interest for achieving high levels of
automation, but also plays a key role in unstructured and dynamic
environments, as in coordinated exploration and mapping
\cite{sal_vision_loc,robust_fusion}. In such scenarios,
fixed infrastructure cannot be used to reduce deployment and management
costs, not to mention potentially cumbersome calibration procedures.

The increasing affordability and availability of sensing technologies
have made robust localisation a fundamental component in the 
development of multi-agent systems.
In this context, methods that impose few system assumptions and 
maintain scalability are especially 
appealing~\cite{low_cost_pf, scalable_CL}, with Ultra-Wideband (UWB) 
based approaches representing a prominent example~\cite{relative_loc}. 
These qualities are essential
in most distributed systems to ensure broad applicability across
different platforms. Moreover, such properties significantly reduce
deployment efforts and simplify setup procedures, which is 
especially desirable in the development of applications designed for
untrained or inexperienced users.

{\textbf{Related work:}} Numerous studies on cooperative
localisation using ranging measurements rely on the availability of
fixed anchor points~\cite{triangulation_UWB, hamer2018self}, commonly referred to as {\em
anchors}, which are fixed reference nodes with known positions used to
guarantee system observability~\cite{SantoroBF21sensorsjournal,
ShamsfakhrPF23measurement}. While effective in controlled
environments, the use of anchors is unsuitable in unstructured and
dynamic scenarios, where they cannot be deployed or maintained. This
is even more evident considering that in the anchor-based systems,
when measurements are not retrieved simultaneously, trilateration usually 
admits multiple solutions~\cite{why3meas}. In~\cite{indistinguishability}, a complete
analysis of the conditions under which a unique trajectory exists when
retrieving sparse distance measurements is established. The results
show that, for any anchor configuration, there always exist
pathological conditions in which no sufficiently high number of
measurements and/or anchors can avoid robot trajectory
indistinguishability, leaving motion control as the only means to 
enforce observability.  While those analyses focus
on anchor-based systems, their results can be applied to multi-agent
scenarios by replacing the anchors with instances of positions of a
second agent, thus showing that cooperative anchor-less localisation
generally admits multiple solutions.

Hence, to guarantee the multi-agent observability, existing solutions
rely on a richer measurement system that provides higher degrees of
information. For example, several works adopt sensors capable of
retrieving both distance and angle
measurements~\cite{valdeira2024maximum, range_bear_CL}. Another family
of approaches imposes stringent requirements on agent motion.  
\rev{For instance, those relying on active motion control include
strategies that maintain a subset of entities stationary while others
are allowed to move~\cite{Amove_Bstationary, whereareyou}}.  In more
general terms, observability-aware formation control has been explored
as a means of overcoming these issues~\cite{observability_aware,
traj_optimization}. This research field is tailored to the development
of control algorithms that generate motion patterns that maximise the
information gained from motion-coupled observations, both to ensure
observability and to improve the estimate accuracy.  While these
methods are proven to be effective, they usually degrade the control
performance, since part of the control effort must be spent to
guarantee the system observability rather than solving the desired
task~\cite{obs_aware_tradeoff}.

{\textbf{Paper contributions:}}
In this work, we propose a robust and scalable algorithm for
multi-agent 2D localisation that operates under minimal
requirements. The method is anchor-less, fully decentralised, and
explicitly accounts for measurement uncertainty. Unlike most existing
approaches, it does not require \rev{external infrastructure, nor} 
control over the agents' motion, making it suitable for heterogeneous 
systems. The algorithm relies only on local
odometry, sparse inter-agent distance measurements (e.g., UWB-based),
and communication with agents within the network, all of which are
realistic and widely available capabilities in distributed robotics.
As numerous studies have shown, under such minimal
requirements, there is no guarantee of system observability.  Rather than imposing observability through
restrictive control assumptions or additional infrastructure, our
approach embraces this limitation by adopting a fundamentally
different philosophy. Instead of aiming for uniqueness, the algorithm
leverages only the available data and maintains the entire set of
feasible solutions.  We consider this paradigm shift to be a key
novelty compared to existing methods.

The proposed solution is a Multi-Hypothesis Bayesian-based
Decentralised Cooperative Localisation (MHDCL) algorithm, inspired by
the Interim Master Decentralised Cooperative Localisation (IMDCL)
framework \cite{IMDCL_first, IMDCL_improved}. The core idea is to
represent the relative poses of agents through multiple
hypotheses, sufficient to cover the space of possible configurations,
and to update this set collaboratively as measurements
arrive. This multi-hypothesis framework offers two main advantages: (i)
it enables localisation of partially connected networks of agents,
whenever conditions permit, by combining information from the entire
group; (ii) it naturally handles the ill-posedness of the problem by
maintaining all feasible solutions consistent with the current
measurement set, thereby representing and quantifying the lack of
information.

Naturally, these benefits come with trade-offs. First, handling a
large number of hypotheses causes significant computational and
storage effort. Second, since the algorithm may output a set of
possible solutions rather than a single estimate, downstream
applications must be designed to accommodate this non-uniqueness.

Regarding practical requirements, the algorithm’s assumptions 
are easily satisfied in real-world applications, as robots are 
typically equipped with odometry or IMU-based feedback loops. 
Inter-agent distances can be measured with radio-based ranging
technologies such as UWB, which often also provide communication
capabilities. 

The rest of the paper is organised as
follows. Section~\ref{sec:background} introduces the problem and
presents the necessary models. Section~\ref{sec:algorithm} provides a
detailed description of the algorithm, while
Section~\ref{sec:practicalities} discusses practical aspects and the
trade-off between computational cost and estimation performance.
Section~\ref{sec:results} presents \pao{experimental } results in realistic
case studies, and Section~\ref{sec:conclusion} concludes the work and
outlines future research directions.

Multimedial material illustrating the proposed approach and its experimental validation is available online.\footnote{Video available at: https://www.youtube.com/watch?v=owNXrBRTyX4}


\section{BACKGROUND AND PROBLEM FORMULATION} \label{sec:background}

\begin{figure}[t]
    \centering
    \includegraphics[width=0.60\linewidth]{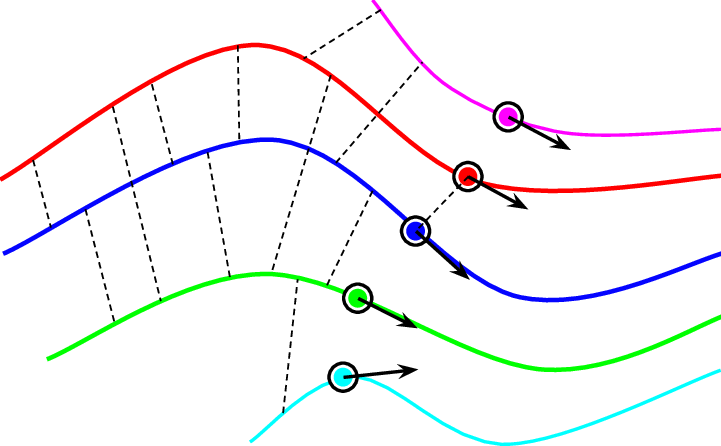}
    \caption{Illustration of the multi-agent system during operation. The agents traverse the environment while collecting inter-agent distance measurements (dashed lines).}
    \label{fig:fleet} 
\end{figure}

A fleet of $n_\text{ag}$ agents is deployed in a 2D
environment as depicted in Figure~\ref{fig:fleet}. Agents are assumed to have access to local odometry
information and to obtain occasional inter-agent distance measurements
through ranging technologies such as UWB. Moreover, we assume that all
agents can communicate with the entire group at any time.  The goal
for each agent running the decentralised algorithm is to estimate the
relative poses of all other agents with respect to its own local
reference frame. 

\subsection{Notation}

Matrices and vectors are denoted as bold letters (e.g., $\mathbf{A}, \mathbf{q}$), whereas scalar quantities and functions use non-bold notation.
The block diagonal matrix of the set $\mathbf{A}_1,..,\mathbf{A}_n$ is written as $\text{diag}(\mathbf{A}_1,..,\mathbf{A}_n)$.
The symbol $\mathbb{R}^n$ denotes the $n$-dimensional Euclidean space. $\mathbf{0}_{n \times m}$ defines the zero matrix of size $n \times m$, and $\mathbf{I}_n$ is the identity matrix of size $n$.
A planar rotation is described by the rotation matrix
\[ \mathbf{R}(\theta)=\begin{bmatrix} \cos(\theta) & -\sin(\theta) & 0 \\ \sin(\theta) & \cos(\theta) & 0 \\  0 & 0 & 1 \end{bmatrix} .\]
Random variables are specified by their distribution. In particular,
$\mathcal{N}(\mu, \sigma^2)$ denotes a Gaussian with mean $\mu$ and variance $\sigma^2$, and $\mathcal{U}(a,b)$ a continuous uniform distribution defined over the interval $[a,b]$.
The expectation of a random variable $\mathbf{x}$ is written as $\mathbb{E}[\mathbf{x}]$.
Estimates are characterised by a state vector $\mathbf{q} \in \mathbb{R}^n$ and its corresponding error covariance $\mathbf{P}\in \mathbb{R}^{n\times n}$. 
Superscripts indicate both the type of estimate and the agent relationship: $p$ and $c$ denote particle and cluster indices, respectively; the first index refers to the agent being estimated, while an optional second index, placed after a comma, denotes the agent maintaining the estimate.
For example, $\mathbf{q}^{pi,j}$ represents the $p$-th particle estimating the state of agent $i$, as maintained by agent $j$.
A superscript “$-$” indicates a propagated (predicted) state.

\subsection{System models}

Each agent $a \in \{1, .., n_\text{ag}\}$ is modelled using a discrete
time model.  Without loss of generality, in this paper, agents are
modelled as unicycles: their poses at discrete time step $k$ are fully
defined by the set of $n=3$ states
$\mathbf{q}_k^a = [x_k^a, y_k^a, \theta_k^a]^\top$, and their dynamics
are described by the following equations of motion
\begin{equation}
    \mathbf{q}_{k+1} = f(\mathbf{q}_k,\mathbf{u}_k) = \begin{bmatrix}
        x_k + \cos(\theta_k)v_k \Delta t \\
        y_k + \sin(\theta_k)v_k \Delta t \\
        \theta_k + \omega_k \Delta t
    \end{bmatrix} ,
    \label{eq:motion_f}
\end{equation}
where
$\mathbf{u}_k=\begin{bmatrix} v_k & \omega_k \end{bmatrix}^{\top} \in
\mathbb{R}^m, \ m=2$, are the noisy model inputs and $\Delta t$ is the
time-step duration. Input uncertainty is assumed to be white,
Gaussian and zero-mean, with covariance matrix
$\mathbf{Q}_k \in \mathbb{R}^{m\times m}$.  Since the motion model is
non-linear, its Jacobians are computed following the concepts defined
by the Extended Kalman Filter (EKF) \cite{konatowski2007comparison}
\begin{equation}
    \begin{aligned}
        \mathbf{A}_k &= \pdv{f(\mathbf{q}_k,\mathbf{u}_k)}{\mathbf{}{q}_k} = \begin{bmatrix} 
            1 & 0 & -\sin(\theta_k) v_k \Delta t \\
            0 & 1 & \cos(\theta_k) v_k \Delta t \\
            0 & 0 & 1
        \end{bmatrix} , \\
        \mathbf{G}_k &= \pdv{f(\mathbf{q}_k,\mathbf{u}_k)}{\mathbf{u}_k} = \begin{bmatrix}
            \cos(\theta_k) \Delta t & 0 \\ \sin(\theta_k) \Delta t & 0 \\ 0 & \Delta t
        \end{bmatrix} .
    \end{aligned}
    \label{eq:motion_jac}
\end{equation}
Ranging measurements between agent $i$ and $j$ at time $k$ are modelled as
$z_{k}^{ij} = h(\mathbf{q}_k^i,\mathbf{q}_k^j) + \epsilon_k^{ij}$ with likelihood function
\begin{equation}
    h(\mathbf{q}_k^i,\mathbf{q}_k^j) = \sqrt{(x_k^i-x_k^j)^2 +
      (y_k^i-y_k^j)^2} .
    \label{eq:model}
\end{equation}
Measurements are assumed to be affected by a zero-mean, white,
Gaussian noise $\epsilon_k^{ij} \sim \mathcal{N}(0, \sigma_m^2)$.
The measurement model is also non-linear and is therefore linearised as follows
\begin{equation}
    \begin{aligned} 
        {\mathbf{H}^{ij}} &= \pdv{h(\mathbf{q}_k^i,\mathbf{q}_k^j)}{(\mathbf{q}^i_k,\mathbf{q}^j_k)}
                               = \frac{1}{h(\mathbf{q}_k^i,\mathbf{q}_k^j)} \begin{bmatrix} d_{xy} & -d_{xy} \end{bmatrix} ,
    \end{aligned}
    \label{eq:range_h}
\end{equation}
where $d_{xy} = \begin{bmatrix} x_k^i-x_k^j & y_k^i-y_k^j & 0  \end{bmatrix}$. \\
To simplify the notation, when one of the two states is null ($\mathbf{0}_{1\times3}$), the null vector is neglected $h(\mathbf{q}_k^i) = \sqrt{x^i_k+y^i_k}$, and its corresponding Jacobian is simply
\begin{equation}
    {\mathbf{H}^i} = \frac{1}{h(\mathbf{q}^i_k)} \begin{bmatrix} x^i_k
      & y^i_k & 0 \end{bmatrix} .
    \label{eq:range_hsingle}
\end{equation}


\section{MHDCL: A MULTI-HYPOTHESIS DECENTRALISED COOPERATIVE
  LOCALISATION \rev{ALGORITHM}} \label{sec:algorithm}

Each agent $a \in \{1,2,..,n_\text{ag}\}$ performs a decentralised
algorithm based on a particle filter to estimate the possible relative
poses of the other agents given sparse distance measurements
$z^{ij}_{k+1}$, obtained between two agents ($i$ and $j$) at time
$k+1$.  Each agent uses its own reference frame for the poses of the
others, so its position is arbitrarily located at the origin, with the heading pointing towards the $x$ axis.
The following section presents a detailed description of the
algorithm, combining theoretical explanations with the corresponding
mathematical formulations.

\subsection{Initialisation} 
Initially, each agent $a \in \{1,2,..,n_\text{ag}\}$ sets its motion vector to zero, providing an initialisation for propagating future motion estimates
\begin{equation}
    \begin{aligned}
        \mathbf{dq}^{a}_{0} = \mathbf{0}_n , \quad
        \mathbf{dP}^{a}_{0} = \mathbf{0}_{n\times n} , \quad \mathbf{d\Phi}_{0}^a = \mathbf{I}_n
    \end{aligned}
    \label{eq:init_motvec}
\end{equation}
These quantities are only known by the agent they represent.

\subsection{Prediction step}
At each time step, every agent $a \in \{1,2,..,n_\text{ag}\}$ uses the equation of motion~\eqref{eq:motion_f}, and its Jacobians~\eqref{eq:motion_jac} to predict its own motion based on odometry data. This prediction updates each agent’s motion vector and the associated uncertainty as follows
\begin{equation}
    \begin{aligned}
      \mathbf{dq}^{a}_{k+1} &= f(\mathbf{dq}^{a}_{k}, \mathbf{u}^{a}_k) , \quad
      \mathbf{d\Phi}^a_{k+1} = \mathbf{A}^a_k \mathbf{d\Phi}^a_{k} , \\
      \mathbf{dP}^{a}_{k+1} &= \mathbf{A}^a_k \mathbf{dP}^{a}_{k}  \mathbf{A}^{a\top}_k + \mathbf{G}_k^a \mathbf{Q}_k \mathbf{G}^{a\top}_k .
    \end{aligned}
\end{equation}

\subsection{Measurement Handling}
At time step $k+1$, if a measurement $z^{ij}_{k+1}$ is available
between agents $i$ and $j$, they either initialise their particle sets
(in the case of the first measurement) or update their existing
estimates. Without loss of generality, the update procedure is
described from the perspective of agent $i$, with agent $j$ considered
as the counterpart. The same steps are simultaneously performed by
agent $j$ with reversed roles, assuming that
$z^{ji}_{k+1} = z^{ij}_{k+1}$.

\subsubsection{Initialisation at first measurement}

If $z^{ij}_{k+1}$ is the first measurement retrieved between agent $i$
and $j$, the receiving agent initialises a set of $n_\text{P}$
particles representing hypotheses for the other agent’s pose,
distributed according to the measurement uncertainty, as described
next:
\begin{algorithmic}[1]
\For{each particle $pj,i \in \{1,..,n_\text{P}\}$}
    \State Draw $\alpha \sim \mathcal{U}(-\pi, \pi)$, $\delta_p \sim \mathcal{N}(0, \sigma_m^2)$
    \State Compute position
    \begin{equation*} \begin{bmatrix} x^{pj,i} \\ y^{pj,i} \end{bmatrix} = (z^{ij}_{k+1} + \delta_p) 
    \begin{bmatrix} \cos(\alpha) \\ \sin(\alpha) \end{bmatrix} \end{equation*}
    \hspace{0.5 cm}
    \State Draw heading: $\theta^{pj,i} \sim \mathcal{U}(-\pi, \pi)$
    \State New particle's pose: $\mathbf{q}^{pj,i} = \begin{bmatrix} x^{pj,i} & y^{pj,i} & \theta^{pj,i} \end{bmatrix}^\top$
    \State Initialise particle covariance matrix:  $\mathbf{P}^{pj,i} = \mathbf{0}_{n\times n}$
\EndFor
\end{algorithmic}

\subsubsection{Update Step at subsequent measurements}
\label{sec:ij_update}
If $z^{ij}_{k+1}$ is not the first measurement between agent
$i$ and $j$, then the particles from both agents can undergo an
update, which eliminates low-probability hypotheses and redistributes particles in regions where the agent $j$ is more likely to be located.
\begin{figure*}[t]
  \centering
  \includegraphics[width=\textwidth]{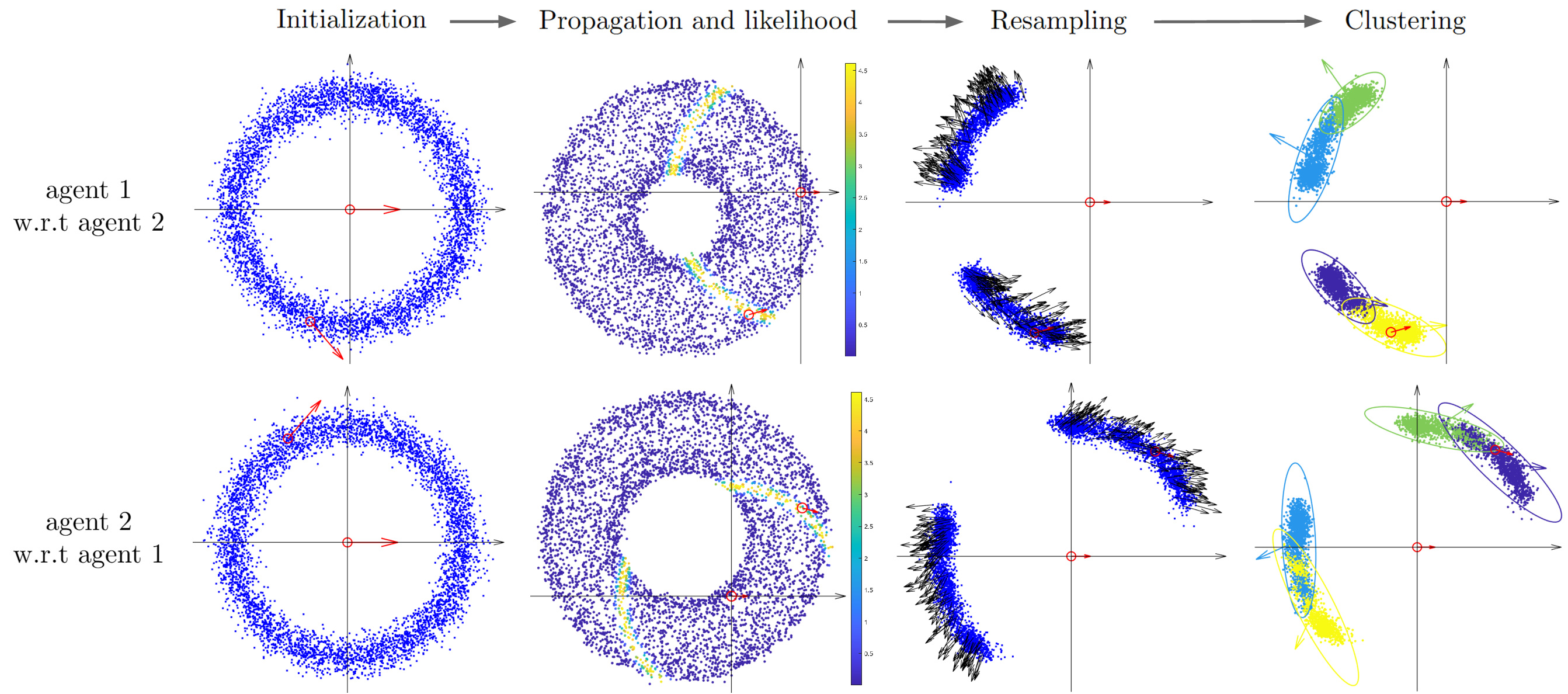}
  \caption{Example of the algorithm’s operation with two distance
    measurements between a pair of agents, leading to initialisation
    at first measurement, then update and clustering at the second
    measurement. Red arrowed circles indicate the true positions of
    the agents.}
  \label{fig:init_updt}
\end{figure*}
Before computing the update, agent $i$ needs to propagate the pose of
each particle of $j$ according to the movements that both agents
performed since the previous measurement was taken. Hence,
agents communicate to each other their motion vector and the
associated uncertainty: for instance,
$(\mathbf{dq}^{j}_{k+1}, \mathbf{d\Phi}^{j}_{k+1} , \mathbf{dP}^{j}_{k+1})$ is sent from $j$ to
$i$, and vice-versa.

First, the pose and the covariance matrix of each particle of agent
$j$ held by the agent $i$ are updated given the $j$-th agent motion
vector $\mathbf{dq}_{k+1}^{j}=[dx^j,dy^j,d\theta^j]^\top$ as
\begin{equation}
    \begin{aligned}
        \tilde{\mathbf{q}}^{pj,i-} &= \mathbf{q}^{pj,i} + \mathbf{R}({\theta^{pj,i}}) \, \mathbf{dq}_{k+1}^{j} , \\
        \tilde{\mathbf{P}}^{pj,i-} &= \widetilde{\mathbf{{d\Phi}}}^{j} \mathbf{P}^{pj,i} \widetilde{\mathbf{{d\Phi}}}^{j\top} + \mathbf{R}({\theta^{pj,i}}) \mathbf{dP}^{j} \mathbf{R}({\theta^{pj,i}})^\top ,
    \end{aligned}
    \label{eq:forward_prop}
\end{equation}
where $\widetilde{\mathbf{{d\Phi}}}^{j} = {\mathbf{R}(\theta^{pj,i})} {\mathbf{{d\Phi}}}^{j} {\mathbf{R}(\theta^{pj,i})}^\top$. \\
Then, their pose is roto-translated according to the $i$-th agent motion
vector $\mathbf{dq}_{k+1}^{i}=[dx^i,dy^i,d\theta^i]^\top$, so that
agent $i$ is always located at the origin of its reference frame, pointing towards the $x$ axis, i.e.
\begin{equation}
    \begin{aligned}
         \mathbf{q}^{pj,i-} &= \mathbf{R}({-d\theta^i}) (\tilde{\mathbf{q}}^{pj,i-} - \mathbf{dq}^{i}) , \\
         \mathbf{P}^{pj,i-} &= \mathbf{R}({-d\theta^i}) (\overline{\mathbf{d\Phi}}^{i}\tilde{\mathbf{P}}^{pj,i-} \overline{\mathbf{d\Phi}}^{i \top} + \overline{\mathbf{dP}}^{i}) \mathbf{R}({-d\theta^i})^\top ,
    \end{aligned} 
    \label{eq:backward_prop}
\end{equation}
where $\overline{\mathbf{M}} = \mathbf{R}(\pi) \mathbf{M} \mathbf{R}(\pi)^\top$, given a generic matrix $\mathbf{M}$.
Once the particles have been propagated, an
update of each hypothesis can be computed, considering the innovation
\begin{equation}
  r^{pj,i} = h(\mathbf{q}^{pj,i-},\mathbf{q}^i) - z^{ij}_{k+1} =
  h(\mathbf{q}^{pj,i-}) - z^{ij}_{k+1} ,
\end{equation}
where $h(\mathbf{q}^{pj,i-})$ is the distance measurement model
defined in~\eqref{eq:model}. Since $h(\mathbf{q}^{pj,i-})$ is
non-linear, the propagation of the prior state uncertainty
$\mathbf{P}^{pj,i-}$ into the measurement space is approximated by
linearising $h$ around the predicted state. The resulting Jacobian
$\mathbf{H}^{pj,i-}$, defined in~\eqref{eq:range_hsingle}, linearly maps pose uncertainty into
measurement uncertainty.

The innovation covariance, which accounts for both the propagated pose
uncertainty and the sensor uncertainty variance $\sigma_m^2$, is given by
\begin{equation}
    \begin{aligned}
        {\sigma_r^{pj,i}}^2 &= \mathbb{E}[{r^{pj,i}}^2] = \mathbb{E}[(h(\mathbf{q}^{pj,i-},\mathbf{q}^i) - z^{ij}_{k+1})^2] \\
        &\approx \mathbf{H}^{pj,i-} \mathbf{P}^{pj,i-} \mathbf{H}^{pj,i-\top} + \sigma_m^2 ,
    \end{aligned}
\end{equation}
which is then used to evaluate the likelihood of the actual
measurement $z^{ij}$ under each particle hypothesis $\mathbf{q}^{pj,i}$, as
\begin{equation}
    \begin{aligned}
        w^{pj,i} &= \mathcal{N}(h(\mathbf{q}^{pj,i-}) \mid z^{ij}, {\sigma_r^{pj,i}}^2) \\
                 &= \frac{1}{\sqrt{2\pi}{\sigma_r^{pj,i}}} \exp\left(-\frac{( h(\mathbf{q}^{pj,i-})-z^{ij})^2}{2 {\sigma_r^{pj,i}}^2} \right) ,
    \end{aligned}
\end{equation}
and, as customary for particle filters, the weights are normalised so
that $\sum_{p=0}^{n_\text{P}} w^{pj,i} = 1$.

In order to remove low-probability hypotheses particles are resampled according to their weight, using the
systematic approach \cite{systematic_resampling}, which is a low-variance method for selecting particles. First, the cumulative
sum of the weights $\mathbf{c} = \{ c_1, .., c_{n_\text{P}}\}$ is
computed, where $c_m = \sum_{p=1}^m w^{pj,i}$. A single random number
is drawn $u_b \sim \mathcal{U}\left(0, {1} / {n_\text{P}}\right)$ to
generate $n_\text{P}$ equally spaced positions
$u_m = u_b + \frac{m-1}{n_{\text{P}}}$
$\forall m = 1, .., n_\text{P}$. For each $u_m$, the smallest
$\bar{m}$ is found such that $c_{\bar{m}} \ge u_m$ and particle
${\mathbf{q}}^{\bar{m}j,i}$ is selected.

For reducing particle degeneracy, after
resampling, a small amount of Gaussian noise is added to each state
component of the resampled particles
\begin{equation}
  \mathbf{q}^{pj,i} = {\mathbf{q}}^{\bar{m}j,i} + \begin{bmatrix}
    \epsilon_x & \epsilon_y & \epsilon_\theta
  \end{bmatrix}^\top ,
\end{equation}
where $\epsilon_x, \epsilon_y \sim \mathcal{N}(0,\sigma_{rxy}^2)$ and
$\epsilon_\theta \sim \mathcal{N}(0,\sigma_{r\theta}^2)$. This
regularisation step promotes more uniform coverage of the state space and
reduces the risk of particle collapse.  The added noise standard
deviations $\sigma_{rxy}, \ \sigma_{r\theta}$ should be kept small
relative to the process dynamics. In fact, excessive noise can
overpower the information gained from measurements, vanishing the
effect of the update.  Resampled particles are assigned with zero
uncertainty: ${\mathbf{P}}^{pj,i} = \mathbf{0}_{n\times n}$.

\subsubsection{Cluster the particles}
After being resampled, particles are clustered using a specialised
Gaussian-von Mises Mixture Model (GVMMM) algorithm, which is a hybrid
mixture model combining Gaussian distributions for describing the
particle spatial arrangement, and von Mises distributions for
describing the heading distribution.  Each cluster represents a
hypothesis, condensing the agglomerated particles into a single
estimate, defined by the mean and covariance, thereby providing a 
compact representation that facilitates communication to other 
agents and reduces computational burdens. 

The result of the GVMMM algorithm is a set of descriptors
$(\mathbf{q}^{cj,i},\mathbf{P}^{cj,i},w^{cj,i},\kappa^{cj,i})$ for
every cluster $c \in \{1,..,n_{cj,i}\}$. The auxiliary descriptor
$w^{cj,i}$ corresponds to the ratio of the cluster, which is a measure
of how many particles it encloses w.r.t. the total number of
particles, and $k^{cj,i}$ is the heading concentration parameter of
the von Mises distribution.
The number of clusters $n_{cj,i}$ necessary to cover all the particles is selected using a Bayesian Information Criterion (BIC).
Figure~\ref{fig:init_updt} illustrates
how the steps described in the previous sections unfold for a pair of
agents.

\subsubsection{Propagation of other agents}
Each time entity $i$ is involved in a measurement, it applies a
propagation of the particles corresponding to all other agents
$a \in \{1,..,n_\text{ag}\}\setminus\{i,j\}$ according to its own
motion $\mathbf{dq}^{i}_{k+1} = [dx^i,dy^i,d\theta^i]^\top$, ensuring that estimates $\mathbf{q}^{pa,i}, \mathbf{P}^{pa,i}$ are referred in the $i$-th agent body-fixed frame, as in~\eqref{eq:backward_prop}.
To account for motion, the clusters are updated by recomputing their descriptors based on the propagated particles.

\subsection{Collaborative updates}

If agents $i$ and $j$ benefited from a measurement update, they can
share their information, allowing other agents to improve their
estimates, even if they were not directly involved in a
measurement. This also enables sparsely connected agents to
reconstruct the poses of the entire fleet. In fact, a well-defined
cluster basically consists of a distance and angle measurement
between agent $i$ and $j$. Given this knowledge, the other agents can
either perform an update of their existing hypotheses on $i$ and $j$,
or, if they are lacking an estimate for just one of the two, they can use it as an initialisation.

\subsubsection{Cluster sharing}
To let the other agents perform their update, agent $i$ broadcasts its
motion vector $(\mathbf{dq}^{i}_{k+1}, \mathbf{dP}_{k+1}^{i})$ and the
descriptors of its clusters
$(\mathbf{q}^{cj,i}, \mathbf{P}^{cj,i},w^{cj,i},\kappa^{cj,i})$, where
$c \in {1,..,n_{cj,i}}$.  In a specular manner, agent $j$ broadcasts its
knowledge.
Given the received motion vectors, agents $a \in \{1,..,n_\text{ag}\} \setminus \{i,j\}$ propagate their
existing agent $i$ particles $\mathbf{q}^{pi,a}, \mathbf{P}^{pi,a}$ as in~\eqref{eq:forward_prop}.
Cluster descriptors are recomputed to account for the propagation, and the procedure is repeated for the agent $j$ particles
and clusters.

\subsubsection{Initialisation at first
  measurement} \label{sec:monte_init} If agent $a$ has never
initialised particles either for agent $i$ or $j$ (not both), then it
can use its preexisting hypotheses and the new $i$-$j$ measurement to
initialise an estimate for the missing agent.  For example, if agent
$a$ has an estimate for agent $i$, but not for $j$, it can use a
Monte-Carlo approach for generating a set of possible poses for agent
$j$, whose algorithmic steps are reported below.
\begin{algorithmic}[1]
\For{each particle $pj,a \in \{1,..,n_\text{P}\}$}
    \State Randomly select one particle ${{p}i,a} \in \{1,..,n_\text{P}\}$ 
    \State Randomly select one cluster ${{c}j,i} \in \{1,..,n_{cj,i}\}$
    \State Generate a pose from the cluster distribution
    \[ \mathbf{q}_{ji} \sim \mathcal{N}(\mathbf{q}^{{c}j,i-},\mathbf{P}^{{c}j,i-}) \]
    \State Compute new particle's pose with regularisation
    \[\mathbf{q}^{pj,a} = {\mathbf{q}}^{{p}i,a-} + \mathbf{R}({{\theta}^{{p}i,a-}}) \mathbf{q}_{ji} + [\epsilon_x,\epsilon_y, \epsilon_\theta ]^\top \] 
    \hspace{0.4 cm} where $\epsilon_x, \epsilon_y \sim \mathcal{N}(0,\sigma^2_{rxy})$ and $\epsilon_\theta \sim \mathcal{N}(0,\sigma^2_{r\theta})$
    \State Initialise particle covariance matrix: $\mathbf{P}^{pj,a} = \mathbf{0}_{n\times n}$
\EndFor
\end{algorithmic}
For this procedure to yield reliable estimates that cover the entire
set of feasible solutions, the prior particles and clusters used in
the initialisation procedure must be sufficiently dense and not
excessively dispersed.

\subsubsection{Update step for other agents} \label{sec:other_upd} If agent $a$ already has an estimate of both agent $i$ and $j$, then it can perform
an update on each set of hypotheses. The update is soft, meaning that
a likelihood is computed as the sum of probabilities of all the
possible combinations. Since the computation of the likelihood for
each particle of one agent against all the particles of the other
agent would be computationally too demanding, particles are checked
against clusters, significantly limiting the number of combinations.

To compute the likelihood between an estimate $\mathbf{q}^{pi,a-}$ of
agent $i$ and a cluster $\mathbf{q}^{cj,a-}$ of agent $j$, given the
measured cluster $\mathbf{q}^{cj,i}$, two innovations are computed:
one for the position and one for the heading, namely
\begin{align}
    r_p &= h(\mathbf{q}^{pi,a-},\mathbf{q}^{cj,a-}) - h(\mathbf{q}^{cj,i}) , \\
    r_h &= (\theta^{cj,a-} - \theta^{pi,a-}) - \theta^{cj,i} .
\end{align}
The innovation is computed as a joint probability combining the consistency of the cluster distance with the two estimates and the agreement between particle and cluster headings. For simplicity, we assume the two likelihoods to be uncorrelated.

The uncertainty of the distance innovation is computed by
linearising the non-linear function as in~\eqref{eq:range_h}
and~\eqref{eq:range_hsingle}
\begin{equation}
    \begin{aligned}
        {\sigma_{rp}}^2 &= \mathbb{E}[(r_p)^2] = \mathbb{E}[(h(\mathbf{q}^{pi,a-},\mathbf{q}^{cj,a-})-h(\mathbf{q}^{cj,i}))^2] \\
        &\approx \mathbf{H}^{picj,a-} \mathbf{P}^{pi,cj-} {\mathbf{H}^{picj,a-}}^\top + \mathbf{H}^{cj,i} \mathbf{P}^{cj,i} {\mathbf{H}^{cj,i}}^\top ,
    \end{aligned}
\end{equation}
where
$\mathbf{P}^{pi,cj-} = \text{diag} \left( \mathbf{P}^{pi,a-}, k_\sigma
  \mathbf{P}^{cj,a-} \right)$, and $k_\sigma$ is tunable
to scale the cluster covariance to properly enclose the set of
hypotheses. Assuming that innovations follow a Gaussian distribution, the likelihood is computed as
\begin{equation}
    \begin{aligned}
        w_{rp} &= \mathcal{N}(h(\mathbf{q}^{pi,a-},\mathbf{q}^{cj,a-}) \mid h(\mathbf{q}^{cj,i}), {\sigma_{rp}}^2) \\
            &= \frac{1}{\sqrt{2\pi}{\sigma_{rp}}} \exp\left( -\frac{ (h(\mathbf{q}^{pi,a-},\mathbf{q}^{cj,a-}) - h(\mathbf{q}^{cj,i}))^2}{2 {\sigma_{rp}}^2} \right) .
        \label{eq:p_innov}
    \end{aligned}
\end{equation}
Instead, the heading innovation is modelled with a von Mises distribution. 
The innovation concentration factor is approximated by combining the
angular dispersions of the three sources using the small-angle
approximation
\(\kappa\approx 1/\sigma^2\)~\cite{mardia2009directional}, hence
\begin{equation} 
    \kappa_{rh} = \left(\frac{1}{\kappa^{cj,i}} + {\sigma_{\theta^{pi,a-}}^2} + \frac{1}{\kappa^{cj,a-}} \right)^{-1} .
\end{equation}
Therefore, the innovation likelihood can be defined as
\begin{equation}
    \begin{aligned}
        w_{rh} &= \text{VM}((\theta^{cj,a-} - \theta^{pi,a-}) \mid
               \theta^{cj,i}, \kappa_{rh})  \\
        &= \frac{1}{2\pi I_0(\kappa_{rh})} \exp \left({\kappa_{rh} \cos( \theta^{cj,a-} - \theta^{pi,a} - \theta^{cj,i})} \right) ,
    \end{aligned}
    \label{eq:h_innov}
\end{equation}
where $I_0(\kappa)$ is the zero order modified Bessel function~\cite{mardia2009directional}.

The compound probability is given by the product of the two
likelihoods: $w_r = w_{rp} w_{rh}$. 
This computation is carried out for every measurement cluster $cj,i \in \{1,..,n_{cj,i}\}$ and for each estimate cluster $ci,a \in \{1,..,n_{ci,a}\}$. The resulting contributions are aggregated by weighted sum with the associated cluster weights $w^{cj,i}$ and $w^{ci,a}$, respectively.

The whole procedure is repeated with the roles of agent $i$
and $j$ reversed. Finally, both sets of particles are redrawn using the
systematic resampling procedure described in
Section~\ref{sec:ij_update}. Each resampled particle inherits the
cluster label of the particle it was drawn from, and the cluster
descriptors are recomputed.

\subsection{Final procedure}
After sharing information with the others, agent $i$ and $j$ reset their
motion vectors as in~\eqref{eq:init_motvec}.  Particles and clusters that did
not undergo an update are carried forward unchanged.


\section{ALGORITHM PRACTICALITIES} \label{sec:practicalities}

Optimisation of computational burden is crucial, as the algorithm is
intended for platforms with limited resources. In this section, we
discuss heuristics and tunable parameters that help mitigate
algorithmic limitations and facilitate deployment on
resource-constrained devices.

\subsection{Lightweight update for other agents}
Due to its combinatorial structure, the update step for other agents described in Sec. \ref{sec:other_upd} is computationally demanding, as it requires evaluating all combinations of particles and clusters, leading to a complexity of $\mathcal{O}(n_\text{P} \cdot n_{cj,a} \cdot n_{cj,i})$. \\
To mitigate this cost, a lightweight approximation of the update is introduced.
Instead of evaluating the likelihood for every particle as in~\eqref{eq:p_innov} and~\eqref{eq:h_innov}, the simplified approach computes the likelihood at the cluster level, resulting in a limited number of combinations $\mathcal{O}(n_{ci,a} \cdot n_{cj,a} \cdot n_{cj,i})$, as follows
\begin{equation}
    \begin{aligned}
        {\sigma_{rp}}^2 &\approx \mathbf{H}^{cicj,a-} \mathbf{P}^{ci,cj-} {\mathbf{H}^{cicj,a-}}^\top + \mathbf{H}^{cj,i} \mathbf{P}^{cj,i} {\mathbf{H}^{cj,i}}^\top , \\
        w_{rp} &= \frac{1}{\sqrt{2\pi}{\sigma_{rp}}} \exp\left( -\frac{ (h(\mathbf{q}^{ci,a-},\mathbf{q}^{cj,a-}) - h(\mathbf{q}^{cj,i}))^2}{2 {\sigma_{rp}}^2} \right) ,
    \end{aligned}
\end{equation}
where
$\mathbf{P}^{ci,cj-} = \text{diag} \left(k_\sigma \mathbf{P}^{ci,a-},
  k_\sigma \mathbf{P}^{cj,a-} \right)$, and
\begin{equation}
    \begin{aligned} 
      \kappa_{rh} &= \left(\frac{1}{\kappa^{cj,i}} + \frac{1}{\kappa^{ci,a}} + \frac{1}{\kappa^{cj,a-}} \right)^{-1} , \\
      w_{rh} &= \frac{1}{2\pi I_0(\kappa_{rh})} \exp \left({\kappa_{rh} \cos( (\theta^{cj,a-} - \theta^{ci,a-}) - \theta^{cj,i})} \right) .
    \end{aligned}
\end{equation}
The resulting cluster-level probabilities are then projected back onto
the particles according to the cluster responsibilities obtained from
the Mixture-Model algorithm. This simplified procedure significantly
reduces computational complexity while preserving the main structure
of the update, at the cost of some resolution.


\subsection{Tunable parameters and limitations}
The performance and scalability of the algorithm are strongly
influenced by the choice of the design parameters. The most relevant
are discussed below.

\paragraph{Number of particles} 
The number of particles $n_\text{P}$ directly controls the granularity
of the representation of the state space. Too few particles may cause
loss of feasible hypotheses, particularly when agents travel long
distances between measurements. This effect is amplified when agents
move fast, when measurements are sparse, or when the number of agents
is large, reducing the effective measurement frequency per agent.

\paragraph{Timeliness of estimates} 
A limitation inherited from~\cite{IMDCL_improved} is the
\textit{staleness} of estimates. Since agents exchange their motion
vectors only when participating in a measurement, the hypotheses of
other agents are estimates of their state at the time of the last
measurement and can quickly become outdated. Increasing measurement
frequency or improving network connectivity can partially mitigate
this issue.

\paragraph{Maximum number of clusters} 
The maximum number of clusters $n_c^{\max}$ controls a trade-off between accuracy and computational cost. The BIC-based selection guarantees the selection of a simple, yet good-fitting model, but as $n_c^{\max}$ increases, so does computation time. In practice, performance does not significantly improve above a certain number of clusters, so $n_c^{\max}$ should be tuned to match the signal-to-noise characteristics of the configuration. For example, when distances are large and uncertainty is low, the annulus defined by the likelihood function is thin and can benefit from a larger number of clusters, while when measurement noise is high relative to distance, a smaller number of clusters is sufficient.


\section{\pao{EXPERIMENTAL VALIDATION}} \label{sec:results}

\begin{figure}[t]
    \centering
    \includegraphics[width=0.8\linewidth]{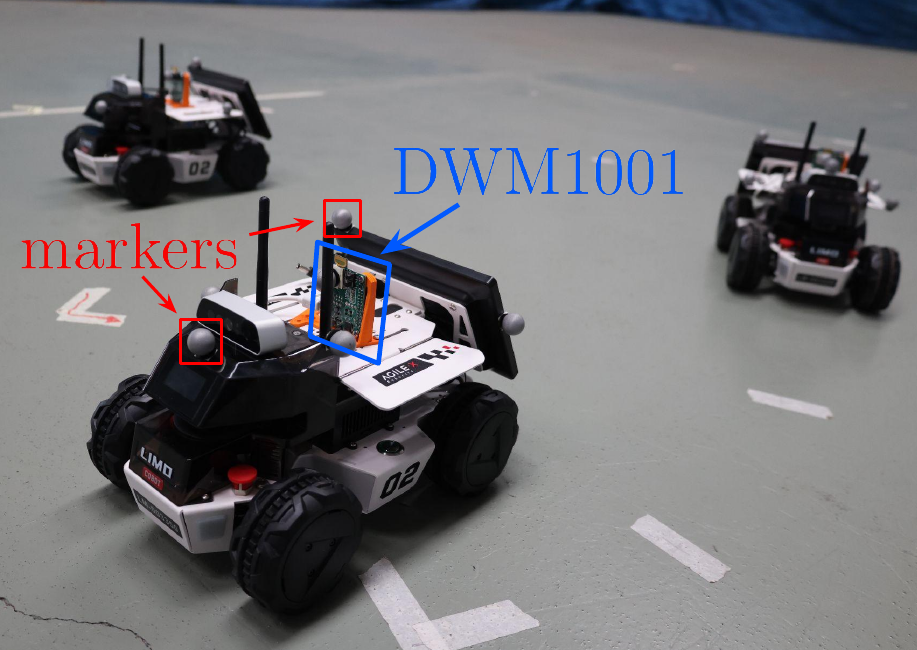}
    \caption{LIMO mobile robots equipped with a DWM1001 UWB module for inter-agent ranging.}
    \label{fig:exp_setup}
\end{figure}

We evaluate the proposed algorithm's in three case studies. Each case highlights different conditions of observability and connectivity \pao{representative of practical deployments}.
The experiments are conducted with multiple LIMO differential-drive mobile robots. As shown in Figure~\ref{fig:exp_setup}, each robot is equipped with a DWM1001 UWB module (Qorvo) for inter-agent ranging and mounts reflective markers enabling tracking by an OptiTrack motion capture system, which provides the ground-truth trajectories for validation. 

The algorithm’s performance is evaluated against this ground truth using three main metrics:
\begin{itemize}
    \item Probability of the real pose: the likelihood that the estimated clusters assign to the true pose of each agent.
    \item Error to the most probable estimate: the distance between the real pose and the most likely cluster.
    \item Cluster area: the total area covered by the clusters at $3\sigma$, used as an indicator of estimate uncertainty.
\end{itemize}
These metrics describe how well the estimates match reality, how
precise the reconstruction is, and how the lack of information evolves over time.

\subsection{Scenario 1 - Weakly observable motion}
\begin{figure}[t]
    \centering
    \includegraphics[width=0.6\linewidth]{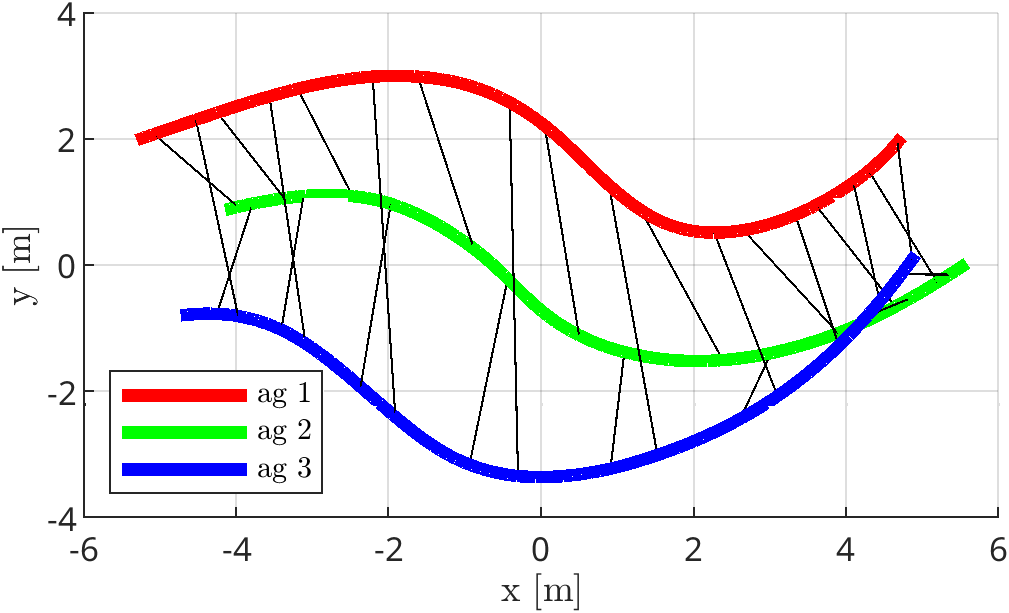}\\
    \vspace{0.1 cm}
    \includegraphics[width=\linewidth,trim=1 1 12 0,clip]{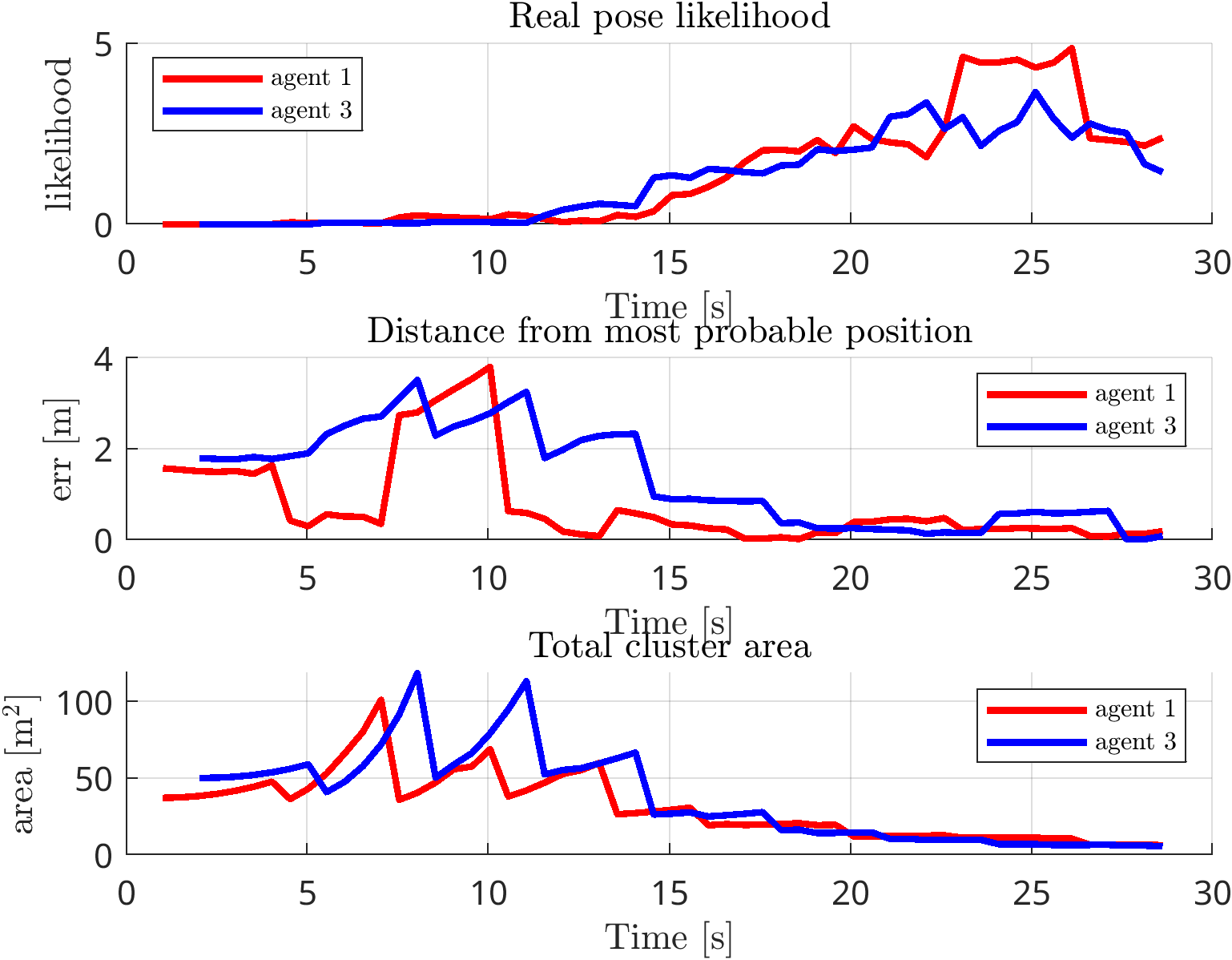}
    \caption{\pao{Experimental } results for Scenario 1. The top panel illustrate
    the ground-truth agent trajectories and inter-agent distance
    measurements, while the bottom panels present the evaluation
    metrics obtained from agent 2.}
    \label{fig:exp1}
\end{figure}

Three agents move along curved paths while keeping roughly constant
relative poses, resembling the behaviour of three individuals walking
together. As can be seen from the results depicted in
Figure~\ref{fig:exp1}, the algorithm successfully reconstructs consistent
estimates despite the measurements being nearly collinear, which are
unobservable configurations~\cite{indistinguishability}.
This experiment highlights the ability of the algorithm to handle
weakly observable configurations, which are typical of human-robot
interaction applications such as coordinated walking or running.

\subsection{Scenario 2: Information transfer}
\begin{figure}[t]
    \centering
    \includegraphics[width=0.6\linewidth]{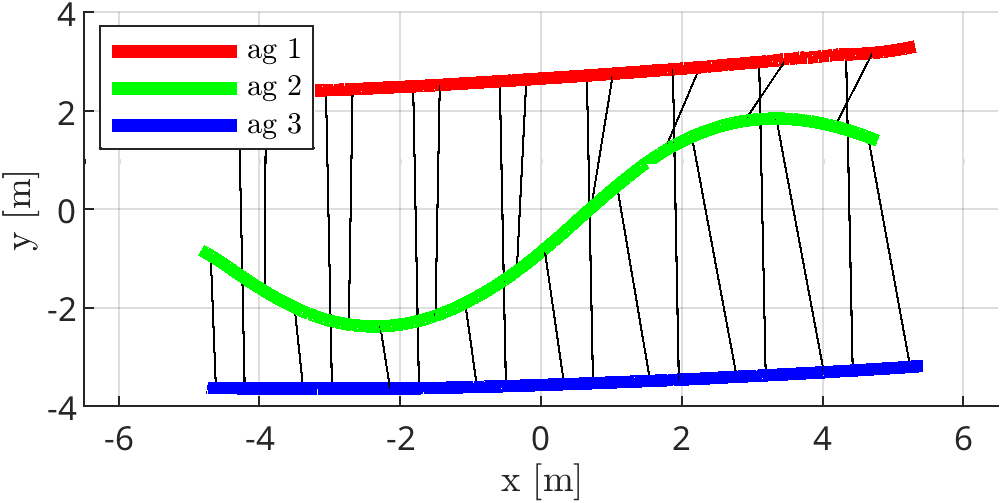}\\
    \vspace{0.1 cm}
    \includegraphics[width=\linewidth,trim=1 1 4 0,clip]{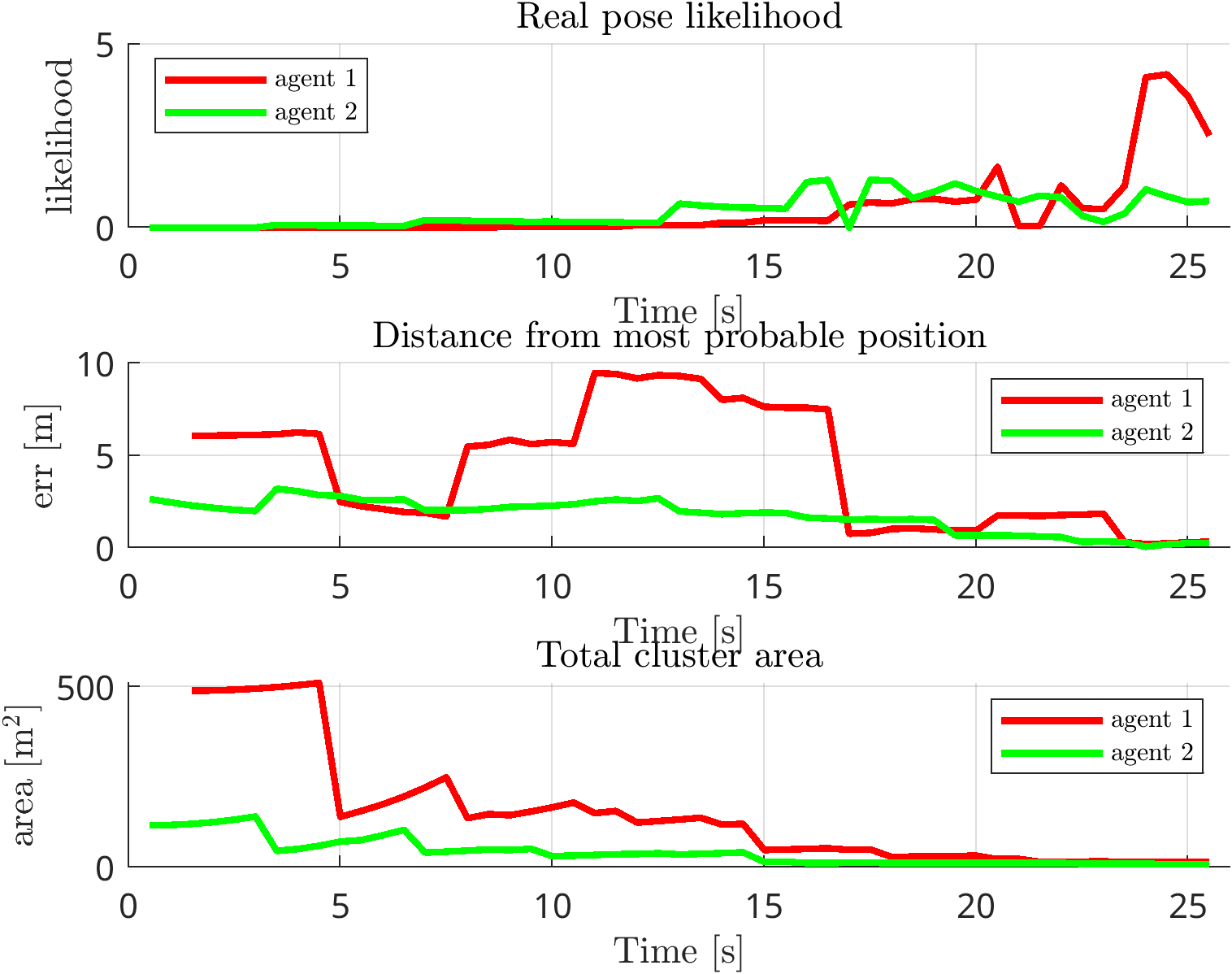}
    \caption{\pao{Experimental } results for Scenario 2. The top and bottom
    panels have the same description as of Figure~\ref{fig:exp1}, and
    the metrics refer to agent 3.}
    \label{fig:exp2}
\end{figure}

In the second scenario, two agents move in parallel trajectories while
a third performs a sinusoidal motion. The third agent trajectory
introduces sufficient diversity in the measurements to resolve the
otherwise unobservable configuration of the two parallel agents. As a
result, accurate estimates are obtained between all pairs of agents
(Figure~\ref{fig:exp2}).
This demonstrates how informative motion from a subset of agents can
propagate through the system, benefiting the localisation of the
entire group.

\subsection{Scenario 3 - Partially connected fleet}
The third scenario considers a fleet of five agents with partial connectivity: agents 1, 2 and 3 are fully connected, agent 4 collects only relative ranging from agent 3, and agent 5 only from agent 4. \pao{While the first two scenarios are validated on real robotic platforms, this larger-scale configuration is evaluated in simulation due to practical considerations.}

\begin{figure}[t]
    \centering
    \includegraphics[width=0.54\linewidth,trim=80 52 55 40,clip]{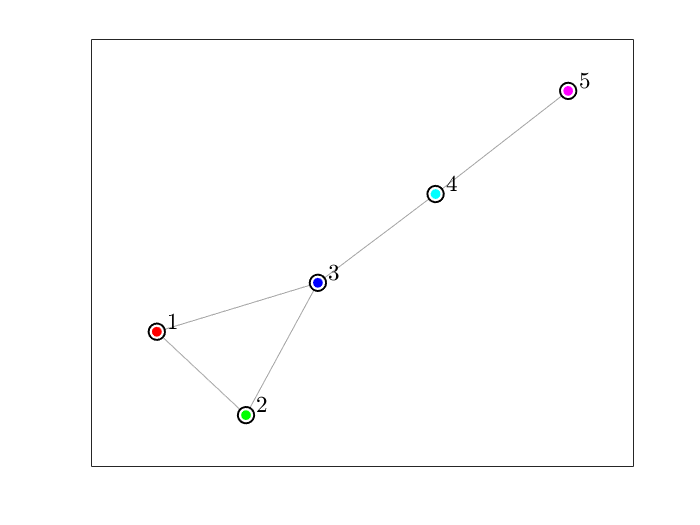}\\
    \vspace{0.1cm}
    \includegraphics[width=\linewidth,trim=20 20 20 16,clip]{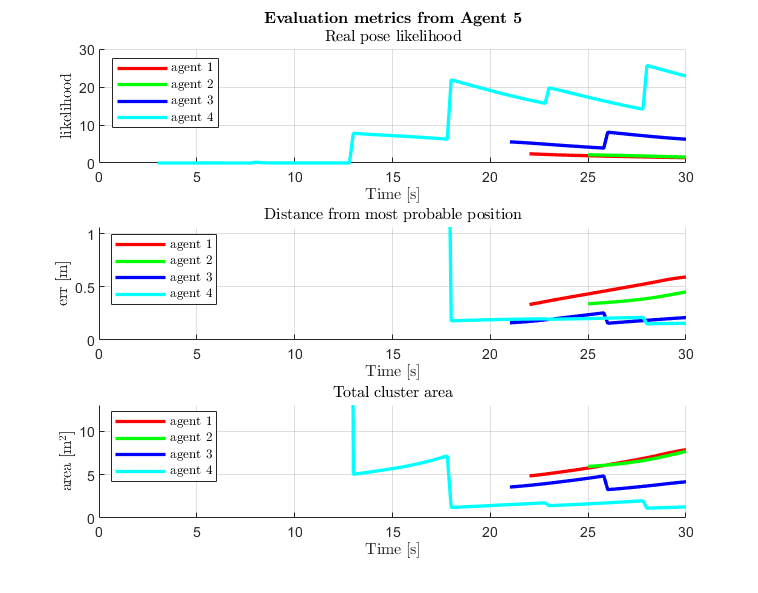}
    \caption{Simulation results for Scenario 3. The top picture depicts
    the network connectivity graph. The bottom panels present
    evaluation metrics obtained from agent 5.}
    \label{fig:exp3}
\end{figure}

Despite the lack of global connectivity, the information gradually propagates across the network, enabling each agent to maintain reliable estimates with all the others. 
Notably, as shown in Figure~\ref{fig:exp3}, even agent 5, which has no direct measurements to agents 1, 2 and 3, is able to estimate their poses with good accuracy.
This result illustrates the scalability of the algorithm to larger, partially connected networks.


\section{CONCLUSIONS} \label{sec:conclusion}

We presented a decentralised cooperative localisation algorithm that
is robust to low-observability conditions by maintaining multiple
hypotheses which capture the full set of feasible solutions. Thanks to
its design, we were able to remove the requirement, common to previous
approaches, of controlling the agent motion to achieve observability,
thus enabling precise task execution without constraining robot behaviour for localisation purposes. Moreover, the method relies on a single sensor per agent and, through information sharing, every new measurement benefits the entire group, enabling agents to improve their estimates collectively.

Through a series of \pao{experiments and simulations}, we demonstrated the
algorithm's ability to preserve accurate estimates even in challenging
scenarios, such as agent trajectories in nearly collinear
configurations. Furthermore, we showed its ability to effectively
propagate information across the group, even if agents are not fully
connected, which is essential for achieving scalability.

Future work will focus on \rev{extending the applicability of the approach to other domains, including multi-rotor aerial robots, as well as humans equipped with wearable devices}. 
\rev{On the algorithmic side, we will explore adaptive particle resampling strategies with noise injection tuned to the estimation uncertainty, as well as optimisations aimed at reducing computational complexity.}





\addtolength{\textheight}{-12cm} 

\bibliographystyle{IEEEtran}
\bibliography{bibliography}

\end{document}